\documentclass[10pt,twocolumn,letterpaper]{article}

 \usepackage{cvpr}              

\usepackage[dvipsnames]{xcolor}

\definecolor{cvprblue}{rgb}{0.21,0.49,0.74}
\usepackage[pagebackref,breaklinks,colorlinks,citecolor=cvprblue]{hyperref}
\usepackage{multirow}

\usepackage{amsmath,amsfonts,bm}

\def\eqref#1{equation~\ref{#1}}

\def\1{\bm{1}}

\def\vx{{\bm{x}}}

\DeclareMathAlphabet{\mathsfit}{\encodingdefault}{\sfdefault}{m}{sl}
\SetMathAlphabet{\mathsfit}{bold}{\encodingdefault}{\sfdefault}{bx}{n}

\def\confName{CVPR}
\def\confYear{2024}

\title{SmoothVideo: Smooth Video Synthesis with Noise Constraints on Diffusion Models for One-shot Video Tuning}

\author{
Liang Peng$^{1, 2}$ 
\quad Haoran Cheng${^{1, 2}}$ 
\quad Zheng Yang${^{2}}$ 
\quad Ruisi Zhao${^{1, 2}}$ 
\quad Linxuan Xia${^{1, 2}}$
\\ 
\quad Chaotian Song${^{1, 2}}$ 
\quad Qinglin Lu${^{3}}$ 
\quad {Boxi Wu${^{1}}$\footnotemark[1]}
\quad Wei Liu${^{3}}$ 
\quad \\
\textsuperscript{\rm 1}Zhejiang University  \quad
\textsuperscript{\rm 2}FABU Inc. \quad
\textsuperscript{\rm 3}Tencent \\
{\tt\small  pengliang@zju.edu.cn \quad wuboxi@zju.edu.cn \quad wl2223@columbia.edu }
}

\begin{document}

\twocolumn[{
\maketitle
\begin{center}
    \captionsetup{type=figure}
    \vspace{-1em}
\newcommand{\imwidth}{1.0\textwidth}
 
\begin{tabular}{@{}c@{}}
\parbox{\imwidth}{\includegraphics[trim=9 9 9 9, clip, width=1.0\linewidth]{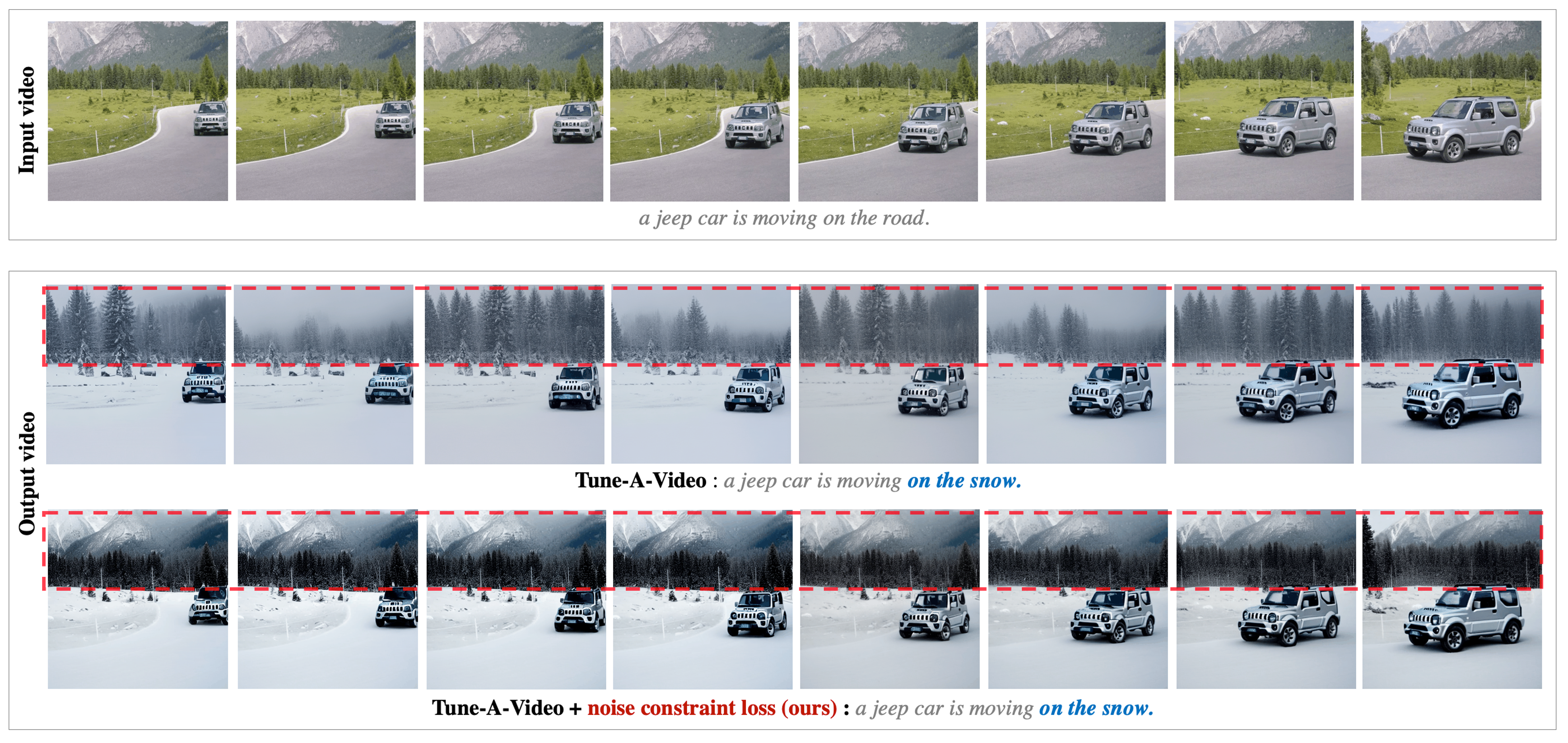}}
\end{tabular}
        \vspace{-4mm}
    \captionof{figure}{
    Comparisons to the baseline.  
    By simply employing the proposed noise constraint loss in the training phase, the model produces smoother videos at the inference stage. 
    We highly recommend the readers refer to the supplementary material for better video comparisons.
    }
    \label{fig:head}
        \vspace{5mm}
\end{center}
}]

\renewcommand{\thefootnote}{\fnsymbol{footnote}} 
\footnotetext[1]{Corresponding authors.} 

\begin{abstract}
	 Significant advancements in generative models have greatly advanced the field of video generation.
	 Recent one-shot video tuning methods, which fine-tune the network on a specific video based on pre-trained text-to-image models (e.g., Stable Diffusion), are popular in the community because of the flexibility.
	However, these methods often produce videos marred by incoherence and inconsistency. 
	To address these limitations, this paper introduces a simple yet effective noise constraint across video frames. 
	This constraint aims to regulate noise predictions across their temporal neighbors, resulting in smooth latents.
	It can be simply included as a loss term during the training phase.
	By applying the loss to existing one-shot video tuning methods, we significantly improve the overall consistency and smoothness of the generated videos. 
	Furthermore, we argue that current video evaluation metrics inadequately capture smoothness. 
	To address this, we introduce a novel metric that considers detailed features and their temporal dynamics.
	Experimental results validate the effectiveness of our approach in producing smoother videos on various one-shot video tuning baselines.
	The source codes and video demos are available at \href{https://github.com/SPengLiang/SmoothVideo}{https://github.com/SPengLiang/SmoothVideo}.
\end{abstract}

\section{Introduction}

	Diffusion-based methods have recently gained widespread acclaim for their excellence in 2D image generation \cite{ho2020denoising,nichol2021improved,song2020denoising,dhariwal2021diffusion,nichol2021glide,ramesh2022hierarchical,saharia2022photorealistic,rombach2022high}. 
	This success has captured the interest and enthusiasm of both the general public and the academic community.
	The generation and editing of videos stand as a fascinating and potentially transformative field, brimming with applications. 
	Despite this, the inherent complexities of video processing have made it a notably challenging endeavor. 	
	Many recent researches \cite{blattmann2023align,singer2022make,wu2023tune,zhao2023controlvideo,guo2023animatediff} focus on this area, yielding substantial advancements.

	This paper focuses on the \textbf{one-shot video tuning} task \cite{wu2023tune}.
	Based on a pre-trained text-to-image (T2I) model (\textit{e.g.}, Stable Diffusion \cite{rombach2022high}), the network incorporates temporal/motion branches and adjustments to attention layers in the model architecture.
	With these modifications, it is capable of fine-tuning the network using an original video and an associated prompt. 	
	This process endows the network with the ability to synthesize videos, enabling the generation of new videos based on new prompts.

	However, existing methods \cite{wu2023tune,zhao2023make,zhao2023controlvideo} often grapple with issues of incoherence and flicker, leading to videos riddled with artifacts and a lack of smoothness. 
	This paper aims to mitigate this problem.
	In the one-shot video tuning task, video frames are derived from initialized latents and noise predictions spanning each timestep in the reverse process. 
	These noise predictions play a crucial role in determining the semantics and spatial arrangement of resulting videos.
	While previous approaches have typically influenced noise predictions using the direct noise regression loss, they often fall short of imposing explicit noise constraints across adjacent video frames. 
	In our analysis of the relationship between noise predictions and corresponding latents (see Section \ref{sec:ana}), we observe that smooth videos and latents tend to exhibit smooth noise predictions.
	Building upon these observations, we introduce a straightforward yet effective method to regulate noise predictions across video frames. 		
	Specifically, we propose a \textbf{noise constraint loss (also referred to as smooth loss)} applied to video noise predictions.
	This loss suppresses the variation in adjacent noise predictions and their associated latents, serving as a regularization term between noise predictions and their temporal neighbors. 
	It facilitates the maintenance of consistent semantics and spatial layout in generated videos, ultimately resulting in smoother latents and videos.

	Furthermore, previous methods  \cite{wu2023tune,zhao2023controlvideo,cong2023flatten,wu2023cvpr}  commonly use CLIP \cite{radford2021learning} frame consistency score to assess frame consistency. 
	This score computes the average cosine similarity between all pairs of video frames.
	However, we identify two significant drawbacks of this metric.
	Firstly, it does not consider the smoothness between adjacent video frames. 
	For instance, a video with frames randomly swapped would yield the same score as the original video, despite the disturbance in the video's temporal order.
	Secondly, CLIP image embeddings used in the metric are coarse-grained, prioritizing overall semantic information while overlooking fine-grained frame details.
	To address these issues, we introduce a new frame consistency metric called video latent score (VL score). 
	This metric considers the smoothness between adjacent frames, replaces CLIP image embeddings with autoencoder \cite{esser2021taming} latents, and incorporates a sliding window design to handle scene motions.
	The VL score provides a more accurate video quality metric in terms of consistency and smoothness.

	Thanks to its simplicity, the proposed method can be seamlessly integrated into existing one-shot video tuning approaches. 
	We have successfully applied our method to enhance three baselines: Tune-A-Video \cite{wu2023tune}, ControlVideo \cite{zhao2023controlvideo}, and Make-A-Protagonist \cite{zhao2023make}.
	The results consistently demonstrate improvements in video smoothness, underscoring the effectiveness of our approach.
	We provide an example in Figure \ref{fig:head}, and we highly recommend that readers refer to the video demos for better comparisons.
	In summary, our contributions are enumerated as follows:

\begin{itemize}
\item
We introduce a simple yet effective method to alleviate the issues of incoherence and flicker in one-shot video tuning.
The method involves a loss term on noise predictions for adjacent video frames, which explicitly regularize the noise predictions.
\item
We emphasize that the previous video evaluation metric, namely CLIP frame consistency score, falls short in precisely reflecting video smoothness. 
Therefore, we propose a new metric called video latent score (VL score), designed to provide a more accurate assessment of the smoothness in synthesized videos.
\item
Our work encompasses extensive experimentation across various baseline models, consistently yielding notable improvements. 
The results demonstrate the effectiveness of the proposed method.
\end{itemize}

\section{Related Work}

 \subsection{Text-to-Image Diffusion Models}
    Text-to-image (T2I) diffusion models \cite{nichol2022glide,ramesh2022hierarchical,saharia2022photorealistic,balaji2022ediffi,ho2022classifier} have attracted much attention in research and industrial communities, owing to the availability of large-scale text-image paired data \cite{schuhmann2022laion} and the power of diffusion models \cite{ho2020denoising,song2020denoising,dhariwal2021diffusion}. 
    Notably, latent diffusion model \cite{rombach2022high}, \textit{i.e.}, Stable Diffusion, proposed to perform the denoising process in an autoencoder’s latent space, effectively reducing computation requirements while retaining the image quality. 
    It is widely employed as a pre-trained model in the community.

\subsection{Text-to-Video Diffusion Models}
    Video diffusion models (VDM) \cite{ho2022video} emulate the achievements of T2I diffusion models by employing a space-time factorized U-Net, integrating training with both image and video datasets. 
    Imagen Video \cite{ho2022imagen} enhances VDM through the application of sequential diffusion models and v-prediction parameterization, enabling the creation of high-resolution videos. 
    Make-A-Video \cite{singer2022make} and MagicVideo \cite{zhou2022magicvideo} are driven by similar goals, focusing on adapting techniques from T2I to text-to-video (T2V)  generation. 
    PYoCo \cite{Ge_2023_ICCV} introduces a novel noise prior method.
    Align-Your-Latents \cite{blattmann2023align} introduces a T2V model which trains separate temporal layers in a T2I model. 
     The field of creating photorealistic and temporally consistent animated frames sequences is emerging. 
    More recently, Animatediff \cite{guo2023animatediff} proposes to insert properly initialized motion modules into the frozen pre-trained text-to-image model and train it on video clips \cite{bain2021frozen}.
    This manner is capable of utilizing various personalized T2I models that derived from the same base T2I models.

\subsection{Text-Driven Video-to-Video Diffusion Models}
	Initial approaches like Text2Live \cite{bar2022text2live} and Gen-1 \cite{esser2023structure} have pioneered layered video representations and text-driven models, yet facing challenges with time-intensive training processes. 
	To address this, subsequent research has shifted towards adapting pre-trained image diffusion models for text-driven video-to-video editing, exemplified by Tune-A-Video \cite{wu2023tune} and its extensions \cite{zhao2023controlvideo,zhao2023make}. 
	These models, including developments like Text2Video-Zero \cite{khachatryan2023text2video}, ControlVideo \cite{zhao2023controlvideo, zhang2023controlvideo}, and FateZero \cite{qi2023fatezero}, aim to preserve structural integrity and improve motion representation, yet often struggle with maintaining visual consistency and detailed texture across frames.
	Further innovations in this domain include zero-shot methods \cite{khachatryan2023text2video,zhang2023controlvideo,qi2023fatezero} that eliminate the need for extensive training phases, utilizing pre-trained diffusion models like InstructPix2Pix \cite{brooks2023instructpix2pix} or ControlNet \cite{zhang2023adding}. 
	These methods, while enhancing flexibility and editing precision, still face challenges in achieving temporal consistency and fine-grained control over the video output. 
	Some recent methods \cite{geyer2023tokenflow,cong2023flatten,ouyang2023codef,yang2023rerender} enforce pixel/feature-level operation to operate video-to-video editing.
	For example, TokenFlow \cite{geyer2023tokenflow} enforces linear combinations between latents based on source correspondences and CoDeF \cite{ouyang2023codef} proposes a canonical content field and a temporal deformation field as a new type of video representation.
	Despite the progress made with techniques like attention map injection and latent feature fusion, maintaining high fidelity in both style and detail during video generation remains an ongoing challenge in this field.
    
    Specifically, the most related method to our work is Tune-A-Video \cite{wu2023tune}, which is the pioneering method of \textbf{one-shot video tuning} task.
    It inflates a pre-trained image diffusion model and finetunes on a specific input video to enable video-to-video editing tasks. 
    Following this line, ControlVideo \cite{zhao2023controlvideo} add image condition for better controlling and Make-A-Protagonist \cite{zhao2023make}  uses textual and visual clues to edit videos to empower individuals to become the protagonists of videos.

\subsection{Evaluation Metrics for Video Synthesis}
    Evaluation metrics are essential for quantitative comparisons.
    Some text-to-video methods \cite{brooks2022generating,singer2022make} employ Fr\'echet Video Distance (FVD) scores \cite{heusel2017gans, unterthiner2018towards} for evaluation.
    Unfortunately, as mentioned in \cite{brooks2022generating},   FVD can be unreliable.
    Most methods \cite{wu2021godiva,wu2022nuwa,singer2022make,blattmann2023align,wu2023tune} employ CLIPSIM (referred to as CLIP score (text alignment)), which calculates the similarities between text and each frame of the video and then take the average value as the semantic matching score.
    This metric evaluates the text alignment between resulting videos and text prompt and is commonly employed.
    To access the frame/temporal consistency, CLIP score (frame consistency) is introduced.
     Some methods \cite{blattmann2023align,qi2023fatezero} compute the cosine similarity of CLIP image emddings between all pairs of consecutive frames and some methods  \cite{wu2023tune,zhao2023controlvideo,cong2023flatten,wu2023cvpr}  compute such embeddings on all frames of output videos and report the average cosine similarity between all pairs of video frames.
    However, this manner of measuring frame consistency has drawbacks as mention above. 
    We propose a new metric to better evaluate the frame consistency and smoothness.

    Interestingly, existing metrics do not precisely align with human preferences.
    As reported in the CVPR 2023 Text Guided Video Editing Competition \cite{wu2023cvpr}, rankings based on current quantitative metrics are often noisy and can even appear random when compared to final human rankings. 
    This suggests that there is room for improvement in the current set of quantitative metrics for video synthesis, and further researches are needed in this direction.

\section{Method}

\subsection{Preliminaries}

\textbf{Denoising diffusion probabilistic models (DDPM).}
DDPM \cite{ho2020denoising} learns the data distribution {\small$q(x_0)$} through Markov's forward and reverse processes. Given the variance schedule $\{\beta_{t}\}_{T}^{t=1}$, noises are gradually added to data $x_0$ in the forward process, and the joint probability of the Markov chain can be described by:
{\small
\begin{equation*}
    q(x_{1:T}) = \prod_{t=1}^{T}q(x_t|x_{t-1})
\end{equation*}
}
where the transition is defined as:
{\small
\begin{equation*}
     q(x_t|x_{t-1}) = \mathcal{N}(x_t; \sqrt{1-\beta_t} x_{t-1}, \beta_t \mathbb{I}), \quad t = 1, \ldots, T.
\end{equation*}
}
Based on Markov property and Bayes' rules, we can express {\small$q(x_t|x_0) = \mathcal{N}(x_t; \sqrt{\Bar{\alpha}_t} x_0, (1 - \Bar{\alpha}_t) \mathbb{I})$}, where {\small $\alpha_t = 1 - \beta_t$} and {\small$\Bar{\alpha}_t = \prod_{s=1}^t \alpha_s$}. And the conditional probabilities {\small$q(x_{t-1}|x_t, x_0)$} can be presented as : 
{\small
\begin{align*}
    q(x_{t-1}|x_t, x_0) & = \mathcal{N}(x_{t-1}; \Tilde{\mu}_t(x_t, x_0), \Tilde{\beta}_t \mathbb{I}), \quad t = 1, \ldots, T, \\
    w.r.t.~~~~ 
    &\Tilde{\mu}_t(x_t, x_0) = \frac{\sqrt{\Bar{\alpha}_t}\beta_t}{1 - \Bar{\alpha}_t} x_0 + \frac{\sqrt{{\alpha}_t} (1 - \Bar{\alpha}_{t-1})}{1 - \Bar{\alpha}_t} x_t.
\end{align*}
}
In the reverse process, DDPM leverage the prior that {\small $p(x_T)$ = $\mathcal{N}(0, \mathbb{I})$} to sample $x1$ \ldots $x_T$ throuth the following transition process:
{\small
\begin{equation*}
    p_\theta(x_{t-1}|x_t) = \mathcal{N}(x_{t-1}; \mu_\theta (x_t, t), \Sigma_\theta(x_t, t)), \quad t = T, \ldots, 1. \\
\end{equation*}
}
 The learnable parameters $\theta$ are trained by maximizing the  variational lower bound of the KL divergence between Gaussian distributions, and the $\Sigma_\theta(x_t, t)$ is set fixed or to be learnable to improve the quality. The improved objective to train model {\small$\epsilon_\theta(x_t, t)$} can be simplified as {\small $\mathbb{E}_{x, \epsilon \sim \mathcal{N}(0, 1), t} \left[ \| \epsilon - \epsilon_\theta(x_t, t) \|^2_2 \right]. $}

~\\
\noindent
\textbf{Denoising diffusion implicit models (DDIM).}
DDIM \cite{song2020denoising} shares the same training strategy as DDPM yet has a non-Markovian forward and and sampling processes. The joint probability of the DDIM forward process is:
{\small
\begin{align*}
     &q(x_t|x_{t-1}) = q(x_t|x_0) \prod_{t=2}^{T}q(x_{t-1}|x_t, x_0)   \\
     &w.r.t.~~~~ x_t = \sqrt{\bar\alpha_t}x_0 +\sqrt{1-\bar\alpha_t}z, z \sim \mathcal{N}(0, \mathbb{I})
\end{align*}
}

$f_\theta(x_{t}, t)$ is the prediction of $x_0$ at $t$ given $x_t$, which can be computed with the trained model $\theta(x_{t}, t)$,:
\begin{align*}\label{eq:f}
  f_\theta(x_{t}, t):= \frac{x_{t} - \sqrt{1-\bar\alpha_t}\epsilon_{\theta}(x_t,t)}{\sqrt{\bar\alpha_{t}}},
 \end{align*}
The corresponding reverse process is as follows:
{\small
\begin{align*}
x_{t-1} = \sqrt{\bar\alpha_{t-1}}f_\theta(x_{t}, t) +  \sqrt{1 - \bar\alpha_{t-1} - \sigma^2_t}{\epsilon}_{\theta}(x_{t}, t) + \sigma^2_t
\end{align*}
}
The sampling process can be controlled by different $\sigma_t$, and by settting the $\sigma_t$ to 0, DDIM's sampling process becomes deterministic, enabling latent variables consistency and fewer sampling steps.

\begin{figure*}[h]
\centering 
		\includegraphics[width=1\linewidth]{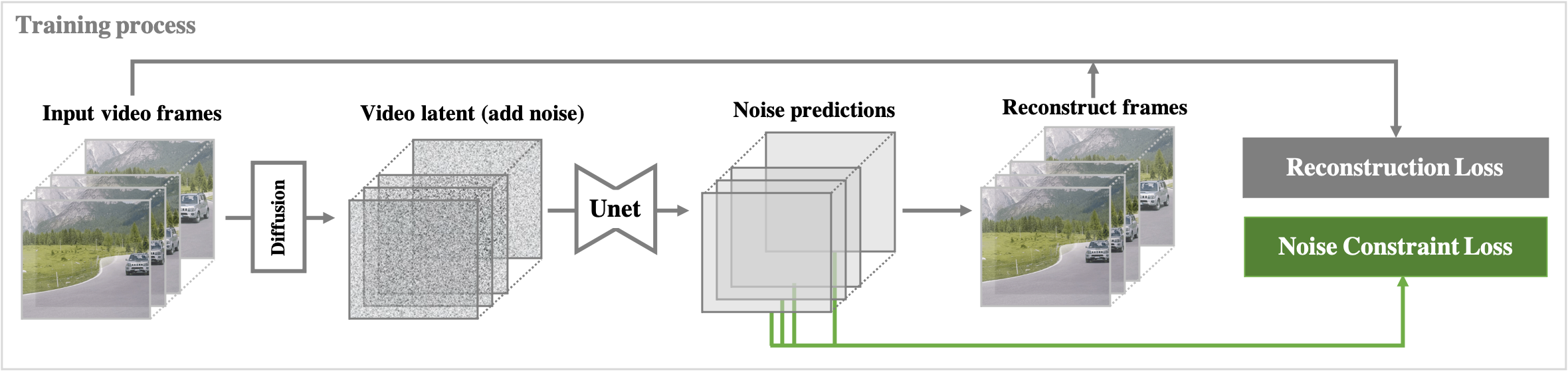}       
		\vspace{-6mm}
		\caption{
				Training overview.
				We apply the proposed noise constraint loss (smooth loss) in the training process for one-shot video tuning. 
				We follow the same pipeline as Tune-A-Video \cite{wu2023tune}, which uses a captioned video to finetune a pre-trained text-to-image (T2I) model with modified network architecture to fit video data.
				}
		\label{fig:framework}
		\vspace{-4mm}
\end{figure*}

\subsection{Problem Definition and Setup}
	Assuming that we have a video prompt and the associated video, the objective of the one-shot video tuning task is to learn customized motion and content within the source video. 	
	Once the model is trained, the tuned diffusion model can then take modified video prompts to synthesize new videos. 
	These new videos share similar motion characteristics with the source video.

\subsection{Analysis on Noise and Latents across Frames}{\label{sec:ana}}
	Smooth videos are characterized by the smoothness of their latent representations across the temporal dimension. 
	These latents are determined by the initialized latents and noise predictions at each reverse timestep. 
	To investigate the connection, we begin by demonstrating the correlation between noise predictions and the resultant latents during the DDIM reverse process.

	To explore the relationship between different video frames, we examine the difference between adjacent latents by applying subtraction in the DDIM reverse process. 
	This can be expressed as:
\begin{gather}
    \vx_{t-1}^{f_{n-1}}  - \vx_{t-1}^{f_{n}} = \frac{\sqrt{\bar\alpha_{t-1}}}{\sqrt{\alpha_{t}}} \left( \vx_{t}^{f_{n-1}}  - \vx_{t}^{f_{n}}  \right) +  \notag \\ 
    \left( \sqrt{1 - \bar\alpha_{t-1}}  -   \frac{\sqrt{\bar\alpha_{t-1}}}{\sqrt{\alpha_{t}}} \sqrt{1 - \bar\alpha_{t}}  \right)  \left(\epsilon_{\vx_t}^{f_{n-1}} - \epsilon_{\vx_t}^{f_{n}} \right)
    \label{eq:e2}
\end{gather}
	where $\vx_{t-1}$ and $\vx_{t} $ are the latents at reverse timestep $t-1$ and $t$, respectively. 
	$f_{n-1}$ and $f_n$ denote the video frame indexes.
	$\epsilon_{\vx_t}$ is the noise prediction at timestep $t$.
	This formula can be re-written as below:
\begin{gather}
\Delta_{\epsilon_{\vx_t}}^{f_{n-1}}  =  C \left( \Delta_{\vx_{t-1}}^{f_{n-1}}-   \frac{\sqrt{\bar\alpha_{t-1}}}{\sqrt{\bar\alpha_{t}}} \Delta_{\vx_t}^{f_{n-1}} \right)      \notag \\
where \quad \Delta_{\epsilon_{\vx_t}}^{f_{n-1}} = \epsilon_{\vx_t}^{f_{n-1}} - \epsilon_{\vx_t}^{f_{n}},     \notag \\ \Delta_{\vx_{t-1}}^{f_{n-1}} =\vx_{t-1}^{f_{n-1}}  - \vx_{t-1}^{f_{n}}, \quad \Delta_{\vx_{t}}^{f_{n-1}} =\vx_{t}^{f_{n-1}}  - \vx_{t}^{f_{n}}    
 \label{eq:e3}
\end{gather}
	In this equation, $C=\frac{\sqrt{\bar\alpha_{t}}}{ \left( \sqrt{\bar\alpha_{t}} \sqrt{1 - \bar\alpha_{t-1}} - \sqrt{\bar\alpha_{t-1}} \sqrt{1 - \bar\alpha_{t}} \right)}$, which is a deterministic value and remains independent of latents and noise predictions. Notably, $\frac{\sqrt{\bar\alpha_{t-1}}}{\sqrt{\bar\alpha_{t}}}$ is approximately equal to 1.
	From this equation, we can discern the relationship between noise and latent differences across frames. 
	Specifically, when latents are smooth (\textit{i.e.}, $\Delta_{\vx_{t-1}}^{f_{n-1}}$ is close to 0), their preceding latents, $\Delta_{\vx_{t}}^{f_{n-1}}$, should also exhibit smoothness. 
	Concurrently, the noise predictions ($\Delta_{\epsilon_{\vx_t}}^{f_{n-1}}$) should be smooth as well, meaning they both approach values close to 0.
	Given that the latents are influenced by noise predictions, we can impose explicit noise constraints between adjacent frames to achieve the goal of producing smooth videos.

\subsection{Noise Constraint Loss}
	Drawing from the above analysis, we advocate for the implementation of the noise constraint loss. 
	It is specifically designed to regulate noise predictions, ensuring smooth noise transition across different frames. 
	We select a random timestep and introduce random noise into the video latents during training, resulting in noisy latents, denoted as $\vx_{t}$. 
	Referencing back to Equation \ref{eq:e3},  a challenge arises in that we cannot retrieve the preceding noisy latents $\vx_{t-1}$, leaving the latent differences $\Delta_{\vx_{t-1}}^{f_{n-1}}$ unknown. 
	Given the gradual nature of the noise addition and denoising process, we postulate that $\Delta_{\vx_{t-1}}^{f_{n-1}}$ closely approximates $\Delta_{\vx_{t}}^{f_{n-1}}$. 
	Therefore, we propose to employ a hyper-parameter to represent this difference. 
	This leads to $\mathbf{L}_{noise-naive} = \lVert \Delta_{\epsilon_{\vx_t}}^{f_{n-1}}  - \frac{C}{\lambda_1} \Delta_{\vx_t}^{f_{n-1}} \rVert$.
	Experimentally, we find that cross-frame differences lead to better performance, as follows:
\begin{gather}
\mathbf{L}_{noise} = \lVert {\Delta^{\prime}}_{\epsilon_{\vx_t}}^{f_{n-1}}  - \frac{C}{\lambda_1} {\Delta^{\prime}}_{\vx_t}^{f_{n-1}} \rVert
 \label{eq:e5}
\end{gather}
	where ${\Delta^{\prime}}_{\epsilon_{\vx_t}}^{f_{n-1}} =  (\epsilon_{\vx_t}^{f_{n-1}} - \epsilon_{\vx_t}^{f_{n}}) + (\epsilon_{\vx_t}^{f_{n-1}} - \epsilon_{\vx_t}^{f_{n-2}})$ and ${\Delta^{\prime}}_{\vx_{t}}^{f_{n-1}} =(\vx_{t}^{f_{n-1}}  - \vx_{t}^{f_{n}}) + (\vx_{t}^{f_{n-1}}  - \vx_{t}^{f_{n-2}})$. 
		Interestingly, during our experimental phase, we attempt to directly regulate the variations in noise predictions, namely, $\mathbf{L}_{noise-simple} = \lVert {\Delta^{\prime}}_{\epsilon_{\vx_t}}^{f_{n-1}}\rVert$.
		This approach also yields promising results.
	This noise constraint loss can be seamlessly interrogated into existing one-shot video tuning methods, by simply including it during the training process.
	The overall loss in the one-shot training process is as follows:
\begin{gather}
\mathbf{L} = \mathbf{L}_{org} + \lambda_2 \mathbf{L}_{noise} 
 \label{eq:e7}
\end{gather}
	where $\mathbf{L}_{org}$ denotes the original mean squared error (MSE) loss associated with noise prediction, while $\lambda_2$ represents the weighting factor for the noise constraint loss (smooth loss). 
	This integration process is visually depicted in Figure \ref{fig:framework}. 
	 To validate the effectiveness of our approach, we conduct experiments on different one-shot video tuning methods. 
	 The experimental results consistently demonstrate improvements on the smoothness of the generated videos.

\subsection{Application on Training-free Methods}
\begin{figure*}[t]
\centering 
		\includegraphics[trim=3 3 3 3, clip, width=1.0\linewidth]{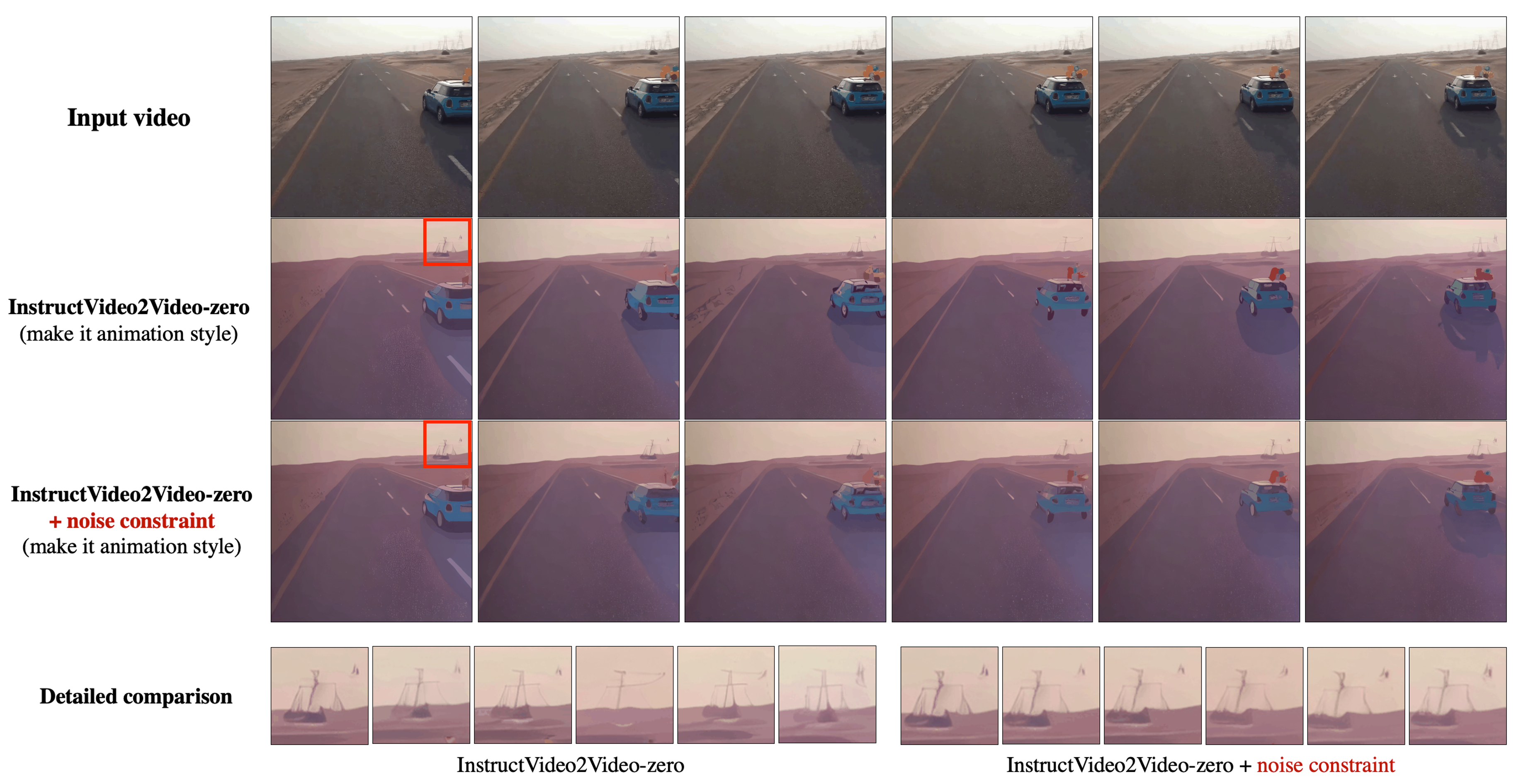}       
		\vspace{-8mm}
		\caption{
			Qualitative comparisons to a training-free baseline.
			Here we compare with the extended video version of InstructPix2Pix \cite{brooks2023instructpix2pix}. 
			We call it InstructVideo2Video-zero.
			We highly recommend the readers refer to the supplementary material for video comparisons. 
			}
		\label{fig:free}
		\vspace{-2mm}
\end{figure*}

	Recent advancements have seen significant success with training-free methodologies \cite{khachatryan2023text2video,qi2023fatezero}. 
	These approaches circumvent the need for retraining networks, opting instead to utilize pre-trained text-to-image/image-to-image models for video editing under text conditions. 
	Let us begin with a simple baseline. 
	Our baseline method uses InstructPix2Pix \cite{brooks2023instructpix2pix} to conduct image-to-image editing, and extend it to achieve video-to-video editing. 
	This naive approach overlooks the crucial aspect of temporal connectivity in videos, leading to the production of incoherent frame sequences. 
	Drawing inspiration from the proposed noise constraint methodology, we adapt our technique by altering the original noise predictions during the inference stage as follows:
\begin{gather}
\epsilon_{\vx_t}^{f_{n-1}}= \epsilon_{\vx_t}^{f_{n-1}} - \lambda_3 ( \Delta_{\epsilon_{\vx_t}}^{f_{n-1}}  - \frac{C}{\lambda_1} \Delta_{\vx_t}^{f_{n-1}} )
 \label{eq:e8}
\end{gather}
	We set $\lambda_3=0.5$.
	The comparative results are presented in Table \ref{tab:free} and Figure \ref{fig:free}. 
	By regulating the noise predictions during the inference stage, we obtain more coherent and smoother results.  	
	Furthermore, we apply this method into previous success training-free methods  \cite{khachatryan2023text2video}  and obtain improvements.
	For more detailed visual comparisons, we invite readers to refer to the video illustrations provided in the supplementary material.

	\begin{figure}[b]
	\vspace{-6mm}
\centering 
		\includegraphics[trim=9 9 9 9, clip, width=0.9\linewidth]{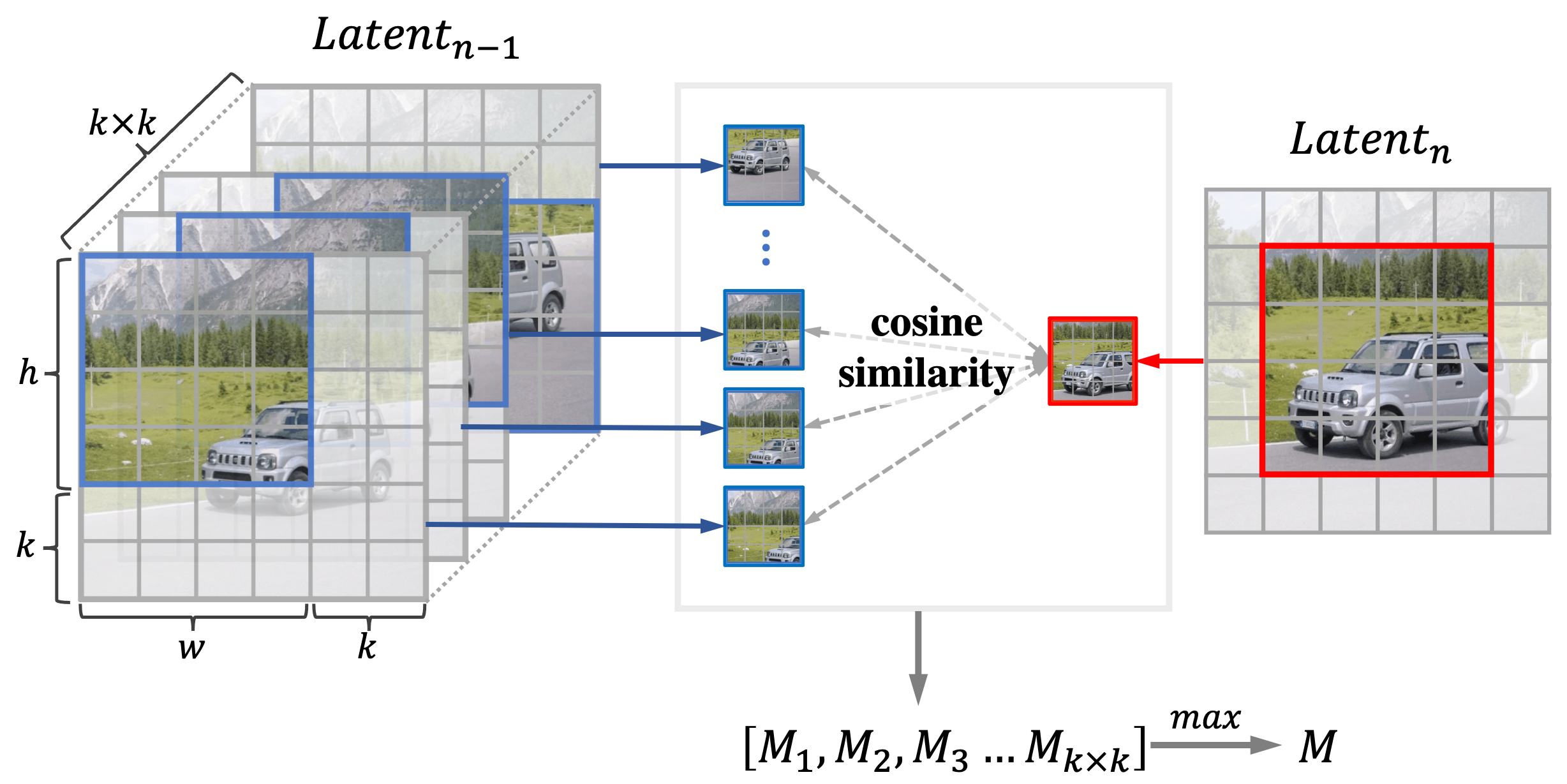}     
		\vspace{-3mm}  
		\caption{
				The computation of video latent score (VL score) between video frame $n$ and the previous frame $n-1$.
				The latents from the previous frame slide with a window of size $h\times w$.
				We calculate the cosine similarity between each resulting latent and the current frame's latent, then select the maximum value to represent the smoothness. 
				This sliding window design is intended to mitigate scene misalignment caused by motion.
			}
		\label{fig:VL}
		\vspace{-2mm}
\end{figure}

\begin{figure*}[t]
\centering 
		\includegraphics[trim=5 5 5 5, clip, width=0.99\linewidth]{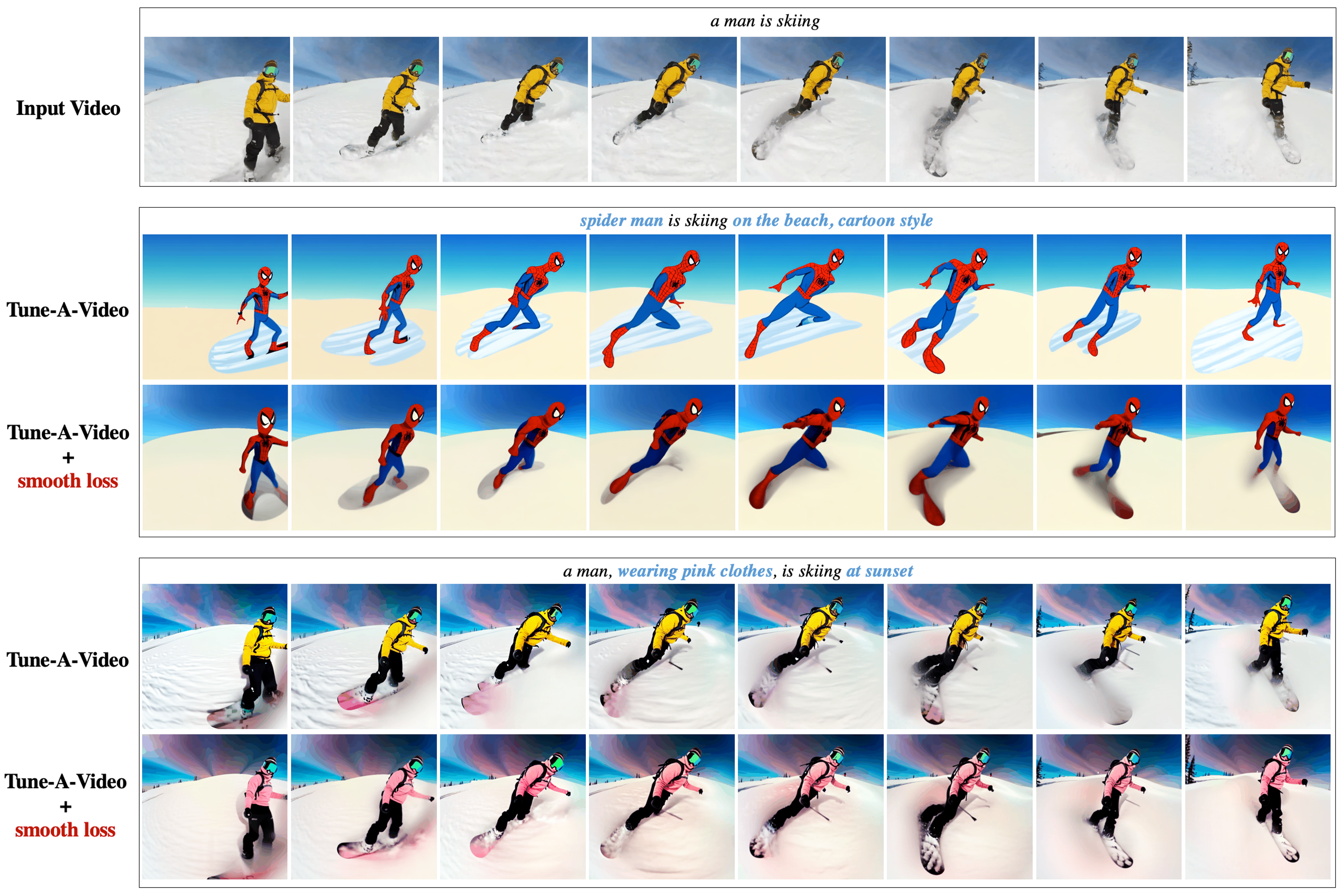}       
		\vspace{-3mm}
		\caption{
				Qualitative comparisons to Tune-A-Video \cite{wu2023tune} baseline.
				Our method significantly improves video consistency and smoothness. 
				For a more detailed and comprehensive comparison, we strongly recommend readers to refer to the supplementary material, which provides additional video comparisons.
			}
		\label{fig:tune}
		\vspace{-3mm}
\end{figure*}

\subsection{Smoothness Metrics for Videos Synthesis}

	Previous methodologies \cite{wu2023tune,zhao2023controlvideo,cong2023flatten,wu2023cvpr} utilize CLIP scores to assess frame consistency. 
	This involves computing CLIP image embeddings for all frames of the output videos and then reporting the average cosine similarity across all pairs of frames. 
	However, we argue that this approach is sub-optimal and present two primary limitations. 
	Firstly, the computing manner on all frame pairs overlooks the crucial aspect of temporal consistency among adjacent frames. 
	This oversight means that videos composed of identical frames but in different orders yield the same score, which is a significant flaw. 
	Secondly, CLIP image embeddings lack the necessary granularity to effectively evaluate video quality. 
	Therefore, they are not ideally suited for assessing the smoothness and consistency across different frames.

	To resolve the above problems, we propose a new metric: video latent score (VL score). 
	This metric involves processing video frames through a latent encoder \cite{esser2021taming} to derive associated latent representations. 
	These latents, with their higher dimensionality (\textit{e.g.}, $4\times64\times64$ for input dimension $512\times512$ compared to $768$ in CLIP \cite{radford2021learning} image embeddings), encapsulate more detailed and informative features crucial for final result assessment. 
	However, simply reshaping and aligning adjacent latents to compute similarity is suboptimal. 
	Adjacent latents often exhibit misalignments, especially when there are scene motions. 
	To address this, as depicted in Figure \ref{fig:VL}, we introduce a sliding window approach. 
	Sliding along with a window, we compute cosine similarities and take the maximum similarity value as the smoothness score. 
	This method is specifically designed to accommodate scene motion dynamics.
	 Formally, this metric is calculated as follows:
\begin{gather}
{\small
{\rm{VL \hspace{1mm} score}} =  \frac{1}{n-1} \sum\limits_{m=1}^{n}  \mathop{{\rm{max}}}\limits_{i,j \in [0, k]}({\rm{Sim}}(\hat{\vx}^{f_m},  \vx^{f_{m-1}}_{i:i+h, j:j+w} ))
 \label{eq:e9}
 }
\end{gather}
	where $\vx$ is the latent, $\hat{\vx}^{f_m} = \vx^{f_m}_{\frac{k}{2}:\frac{k}{2}+h, \frac{k}{2}:\frac{k}{2}+w}$, and $h,w$ denotes the sliding window size.
	$h+k$ and $w+k$ are the latent height and width, respectively.
	We set $k=8$. 
	$\rm{Sim}$ refers to reshaping the latents and calculating the cosine similarity.

\begin{figure*}[t]
\centering 
		\includegraphics[trim=3 3 3 3, clip, width=1.0\linewidth]{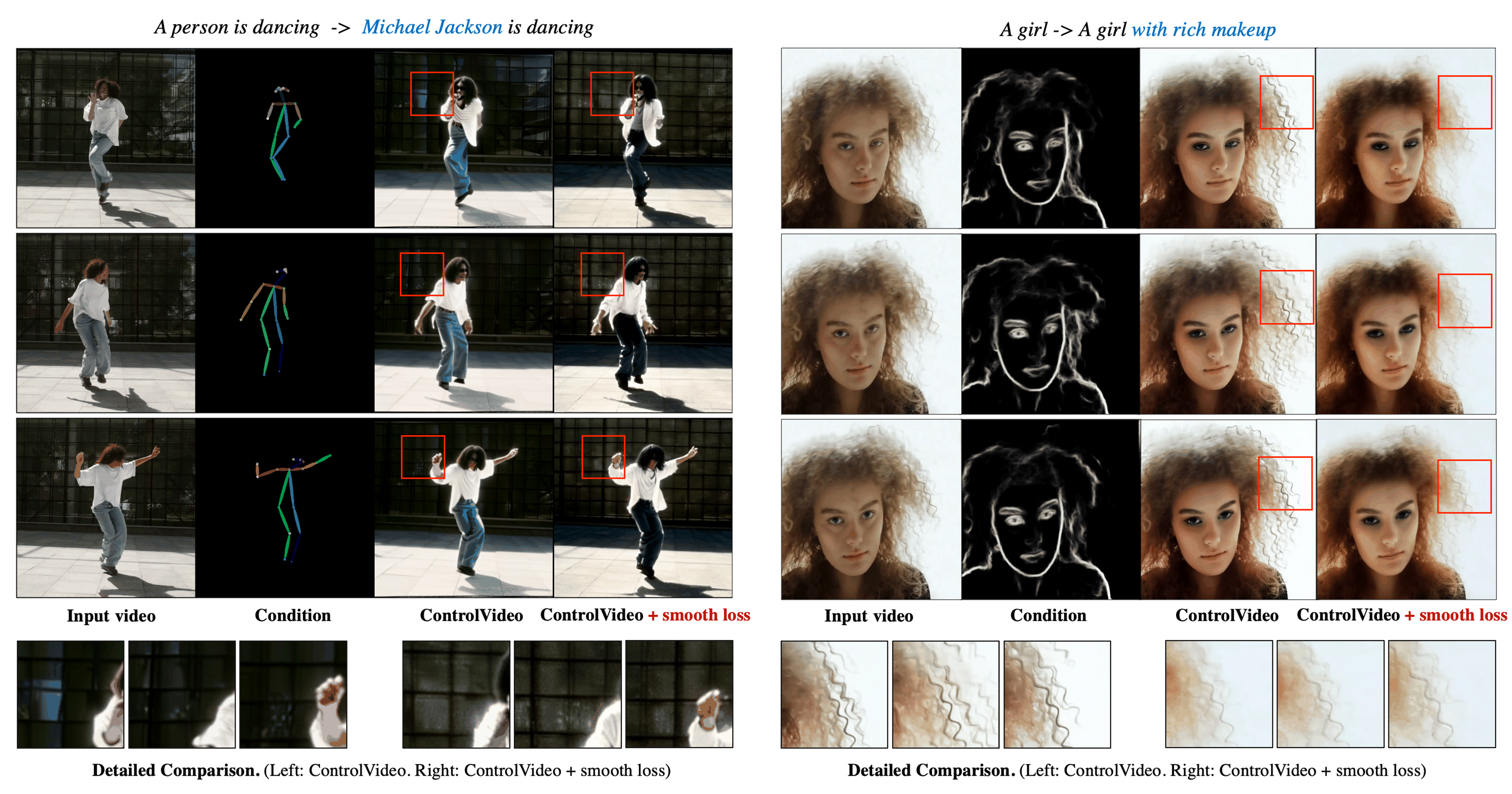}       
		\vspace{-6mm}
		\caption{
				Qualitative comparisons to ControlVideo \cite{zhao2023controlvideo} baseline.
				More detailed and comprehensive video comparisons are included in the supplementary material.
			}
		\label{fig:controlvideo}
		\vspace{-2mm}
\end{figure*}

\begin{figure}[t]
\centering 
		\includegraphics[trim=3 3 3 3, clip, width=1.0\linewidth]{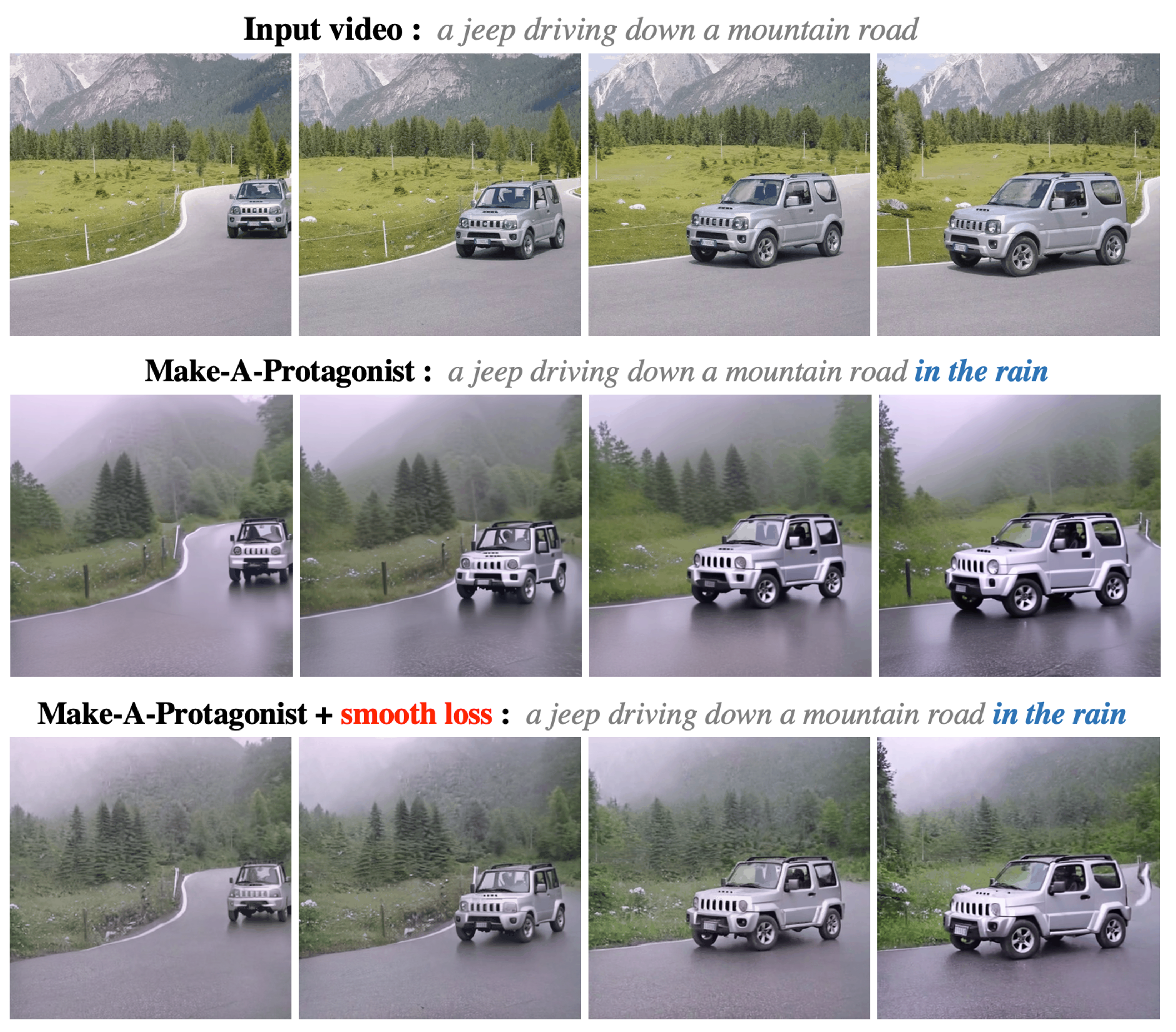}  
		\vspace{-7mm}     
		\caption{
				Qualitative comparisons to Make-A-Protagonist \cite{zhao2023make} baseline.
			}
		\label{fig:make}
		\vspace{-5mm}
\end{figure}

\section{Experiments}

\subsection{Implementation Details}
Our implementation is based on official released codes of each baseline.
For each baseline, we follow its data preprocess strategy (\textit{e.g.}, resolution, frame rate, and frames) and follow the same training scheme (\textit{e.g.}, learning rate, training steps, batch size).
We set $\lambda_1$ in Equation \ref{eq:e5} to 1000, and the smooth loss weight $\lambda_2$ in Equation \ref{eq:e7} to 0.2.
At the inference stage, we use DDIM sampler with classifier-free guidance. 
For the one-shot video tuning task, following previous works, we use the DDIM inversion latents as the initial latents.
For Tune-A-Video \cite{wu2023tune} baseline, we modify its original guidance from 12.5 to 7.5 at the inference stage.
It is because the original value is too high and tends to produce severe flicker.
For the training-free methods, we initialize latents with the same seed across video frames.
All experiments are conducted on a NVIDIA A100 GPU.

\subsection{Dataset and Metrics}
	For Tune-A-Video \cite{wu2023tune} baseline, we employ LOVEU-TGVE-2023 dataset \cite{wu2023cvpr}, which consists of 76 videos with 304 edited prompts. 
	Specifically, it contains 16 videos from DAVIS dataset \cite{pont20172017}, 37 videos from Videvo, 23 videos from YouTube.
	Each video has 4 editing prompts, and all videos are creative commons licensed. 
	Each video consists of either 32 or 128 frames, with a resolution of 480$\times$480.
	We also use additional videos from DAVIS dataset for visualization.
	For metrics, we employ CLIP text alignment score, CLIP frame consistency score, and the propose VL score for quantitive evaluation.
	For ControlVideo \cite{zhao2023controlvideo}, Make-A-Protagonist \cite{zhao2023make}, and training-free baselines \cite{brooks2023instructpix2pix,khachatryan2023text2video}, we use the provided videos in their official implementation for evaluation.
	Such videos mainly come from DAVIS dataset.

\begin{table}[h]
\resizebox{0.5\textwidth}{!}{%
\begin{tabular}{lccc}
\toprule
  Method & CLIP score (T) & CLIP score (F) & VL score  \\
  \midrule
   Tune-A-Video \cite{wu2023tune} & \bf 26.01 & 93.16 & 79.36 \\
      Tune-A-Video \cite{wu2023tune} +smooth loss& 25.22 & \bf 93.54 &\bf 82.10  \\
  \midrule
Make-A-Protagonist \cite{zhao2023make} &\bf 29.18 & 92.98 & 64.29  \\
Make-A-Protagonist \cite{zhao2023make}+smooth loss& 28.89 &\bf  93.33 & \bf 67.77 \\
\midrule
ControlVideo \cite{zhao2023controlvideo} & 20.17 & 93.12 & 77.36  \\
ControlVideo \cite{zhao2023controlvideo}+smooth loss& 20.17 & 93.12 & \bf 79.28  \\
  \bottomrule
\end{tabular}
}
\vspace{-3mm}
\caption{
Quantitative comparisons to one-shot video tuning baselines. 
CLIP score (T) refers to CLIP text alignment score and CLIP score (F) is CLIP frame consistency score.
 VL score denotes the proposed metric.
 Our method boosts the smoothness metrics with significant margins.
Please note that for ControlVideo \cite{zhao2023controlvideo} baseline, we have the same CLIP score (T) and CLIP score (F), but higher VL score.
Our manual visual examination validate the higher smoothness of our method.
We suggest readers refer to the supplementary material for better comparisons.
}
\label{tab:one-shot}
\vspace{-2mm}
\end{table}

\subsection{Comparisons to Baselines}
We apply our method on different baselines. 

\textbf{1) Tune-A-Video} \cite{wu2023tune}.
 It is a pioneering one-shot video tuning method.
 Table \ref{tab:one-shot} provides quantitative comparisons.
 After employing smooth loss, the baseline obtains significant improvements over smoothness metrics.
 We also note that our method downgrades the text alignment score.
The smooth loss plays a trade-off role between text alignment and smoothness because it affects the regular training process.
 We show qualitative comparisons in Figure \ref{fig:tune}. 
 It can be easily seen that our method performs better.

\textbf{2) Make-A-Protagonist} \cite{zhao2023make}.
It is a one-shot video tuning method that can replace the video protagonist with tailored role.
Table \ref{tab:one-shot} provides quantitative comparisons and Figure \ref{fig:make} provides qualitative comparisons.
Our method exhibits better temporal and frame consistency.

\textbf{3) ControlVideo} \cite{zhao2023controlvideo}.
It is a one-shot video tuning method that is capable of combing with different image conditions (\textit{e.g.}, pose, edge, and depth maps) for controlling.
Table \ref{tab:one-shot} provides quantitative comparisons and Figure \ref{fig:controlvideo} provides qualitative comparisons.
Interestingly, we observe that both text alignment scores and clip frame consistency scores are the same, but VL scores are much different.
By visually evaluate the results, our method indeed performs better.
It indicates that the proposed VL score is able to more accurately reflect the human preference.

\textbf{4) Training-free methods.} 
	We extend the noise constraint on a simple training-free method that extends InstructPix2Pix \cite{brooks2023instructpix2pix} to the video area, called InstructVideo2Video-zero.
	We also provide a comparison on a strong baseline: Text2Video-Zero (Video InstructPix2Pix setting) \cite{khachatryan2023text2video}.
	Even without training, the proposed noise constraint consistently improves the results, as shown in Table \ref{tab:free} and Figure \ref{fig:free}.

\begin{table}[t]
\resizebox{0.5\textwidth}{!}{%
\begin{tabular}{lccc}
\toprule
  Method & CLIP score (T) & CLIP score (F) & VL score  \\
  \midrule
InstructVideo2Video-zero\cite{brooks2023instructpix2pix} & 24.59 & 92.95 & 85.88    \\
InstructVideo2Video-zero\cite{brooks2023instructpix2pix}+noise constraint& \bf 25.26 & \bf 93.65 &\bf 86.16   \\ 
\midrule
Video InstructPix2Pix\cite{khachatryan2023text2video}   & 25.59 & 94.84 & 86.25   \\ 
Video InstructPix2Pix \cite{khachatryan2023text2video}+noise constraint &\bf 26.67 &\bf 94.77 &\bf 86.61   \\ 
  \bottomrule
\end{tabular}
}
\vspace{-3mm}
\caption{
Quantitative comparisons on training-free methods.
CLIP score (T) refers to CLIP text alignment score and CLIP score (F) is CLIP frame consistency score.
 VL score denotes the proposed metric.
 InstructVideo2Video-zero refers to the extended video version on InstructPix2Pix \cite{brooks2023instructpix2pix}. 
 Video InstructPix2Pix is the video-to-video setting in Text2Video-Zero \cite{khachatryan2023text2video}.
}
\label{tab:free}
\vspace{-2mm}
\end{table}

\section{Limitations}
	The proposed method has some limitations.
	The noise constraint loss acts like a regular term in the final loss.
	It may downgrade the text alignment in some cases.
	For the sliding window design in the proposed VL score metric, it cannot deal with scene motions like zoom in and zoom out.
	We encourage future work to address these problems.

\section{Conclusion}
	In this paper, we analyze the latent and noise relationship across video frames.
	We find that smooth noise predictions facilitate smooth video synthesis for the one-shot video tuning task.
	Based on this, we introduce a noise constraint loss (smooth loss), to regulate noise predictions  across adjacent video frames.
	The proposed loss can be easily applied into existing one-shot video tuning methods.
	We also extend the noise constraint into training-free methods and obtain improvements.
	Furthermore, we argue that previous metrics on video smoothness has drawbacks and introduce a new metric: VL score.
	It considers the adjacent frame smoothness using fine-grained features, meanwhile alleviating the impact of scene movements.
	Experiments demonstrate the effectiveness of our method.

{
    \small
    \bibliographystyle{ieeenat_fullname}
    \bibliography{main}
}

\end{document}